\documentclass{llncs}
\usepackage{amsmath}
\usepackage{array}
\usepackage{booktabs} 
\usepackage{graphicx}
\usepackage{makeidx}  
\usepackage{placeins}
\usepackage{subcaption}
\newcolumntype{L}[1]{>{\raggedright\let\newline\\\arraybackslash\hspace{0pt}}m{#1}}
\newcolumntype{C}[1]{>{\centering\let\newline\\\arraybackslash\hspace{0pt}}m{#1}}
\newcolumntype{R}[1]{>{\raggedleft\let\newline\\\arraybackslash\hspace{0pt}}m{#1}}
\begin{document}

\newcommand\blfootnote[1]{%
  \begingroup
  \renewcommand\thefootnote{}\footnote{#1}%
  \addtocounter{footnote}{-1}%
  \endgroup
}

\mainmatter              
\title{3-D Convolutional Neural Networks for Glioblastoma Segmentation}
\titlerunning{3DConvNet}  
%
\author{Darvin Yi$^{1,\star}$ \and Mu Zhou$^{1,\star}$ \and Zhao Chen\inst{2} \and Olivier Gevaert\inst{1}}
\authorrunning{Darvin Yi et al.} 
%
\tocauthor{Darvin Yi, Zhao Chen, Mu Zhou, Olivier Gevaert}
\institute{Stanford Center for Biomedical Informatics Research, Stanford University, CA\\
\and
Department of Physics, Stanford University, CA\\
}

\maketitle              

\begin{abstract}
Convolutional Nerual Networks (CNN) have emerged as powerful tools for learning discriminative image features. In this paper, we propose a framework of 3-D fully CNN models for Glioblastoma segmentation from multi-modality MRI data. By generalizing CNN models to true 3-D convolutions in learning 3-D tumor MRI data, the proposed approach utilizes a unique network architecture to decouple image pixels. Specifically, we design a convolutional layer with pre-defined Difference-of-Gaussian (DoG) filters to perform true 3-D convolution incorporating local neighborhood information at each pixel. We then use three trained convolutional layers that act to decouple voxels from the initial 3-D convolution. The proposed framework allows identification of high-level tumor structures on MRI. We evaluate segmentation performance on the BRATS segmentation dataset with 274 tumor samples. Extensive experimental results demonstrate encouraging performance of the proposed approach comparing to the state-of-the-art methods. Our data-driven approach achieves a median Dice score accuracy of 89\% in whole tumor glioblastoma segmentation, revealing a generalized low-bias possibility to learn from medium-size MRI datasets.

\end{abstract}


\blfootnote{$^{\star}$These two authors contributed equally.}

\section{Introduction}
Glioblastoma (GBM) is a highly malignant brain tumor with a dismal prognosis \cite{Adamson09}. Most patients experience disease progression within 7-10 months, and targeted therapies have not increased survival \cite{Omuro13,Omuro07}. Accurate brain tumor segmentation is a significant yet challenging task for follow-up computer-aided diagnosis. Semi-automatic segmentation remains a bottleneck in mining medical imaging data with a lack of definitive guidance of human experts involvement. Automated methods such as graph cuts method tend to lead high-bias models that have not significantly improved accuracy \cite{Njeh15}. \\
%
%
\indent While data-driven models like Convolutional Neural Networks (CNNs) are increasingly prevalent \cite{Deng09,Krizhevsky12,Lawrence97,Wei}, high variance limits their use for medical image analysis as many medical data sets have at most hundreds of patient samples. Strategies like data augmentation and transfer learning may gain success in creating more generalizable low-bias models \cite{Cui14,Yosinski14}, but they do not consider the development of network structures incorporating specific domain knowledge in tumor MRI. In this paper, we identify two key weaknesses in previous approaches by applying CNN models to medium sized imaging data sets. Canonically, CNNs utilize layers of 2-D convolutions as filters for feature learning, feeding the outputs of these convolutional layers into a fully connected neural network. We propose a 3-D CNN model for brain tumor segmentation by generalizing the conventional 2-D architecture to fully take advantage of 3-D multi-modality MRI data. In addition, we propose several important advances leading to accurate segmentation performance. \\ 
\indent First, most prior methods for volumetric image data use either 2-D convolutions or limited 3-D convolutions on the $xy$, $xz$, and $yz$ planes. By contrast, we propose a true generalization to 3-D CNNs, made it computationally feasible by the transformation into Fourier space. Such innovation renders our system more robust with minimal loss of spatial information during convolution. \\
\indent Second, the use of CNN models with medium data sets likely lead to high variance due to the lack of training data for learn network weights. Related algorithms have been proposed to use pre-trained filters from ImageNet \cite{Deng09}, but these 2-D filters maximize object classification from real-world images rather than volumetric medical images \cite{Deng09,Krizhevsky12,Bar15}. Since texture filters have been proven as effective tools for image data analysis, in this study, we perform 3-D convolutions using pre-defined difference of Gaussian (DoG) filters, which are rotationally symmetric and act as effective blob detectors \cite{sift}.  Subsequent CNN layers use $1 \times 1 \times 1$ convolutions to decouple pixels, expanding effective data size from patient number to total pixel count and significantly reducing variance. Our 3-D CNN leverages the structure of medical imaging data to train a robust and efficient algorithm for learning from 3-D images. We apply our framework for segmenting brain tumors and compare with previous approaches as well as results provided by expert annotations.


\section{Methods}
\subsection{3-D Convolutional Neural Networks Architecture}
\begin{figure}[htb]
	\centering
	\includegraphics[width=0.9\textwidth]{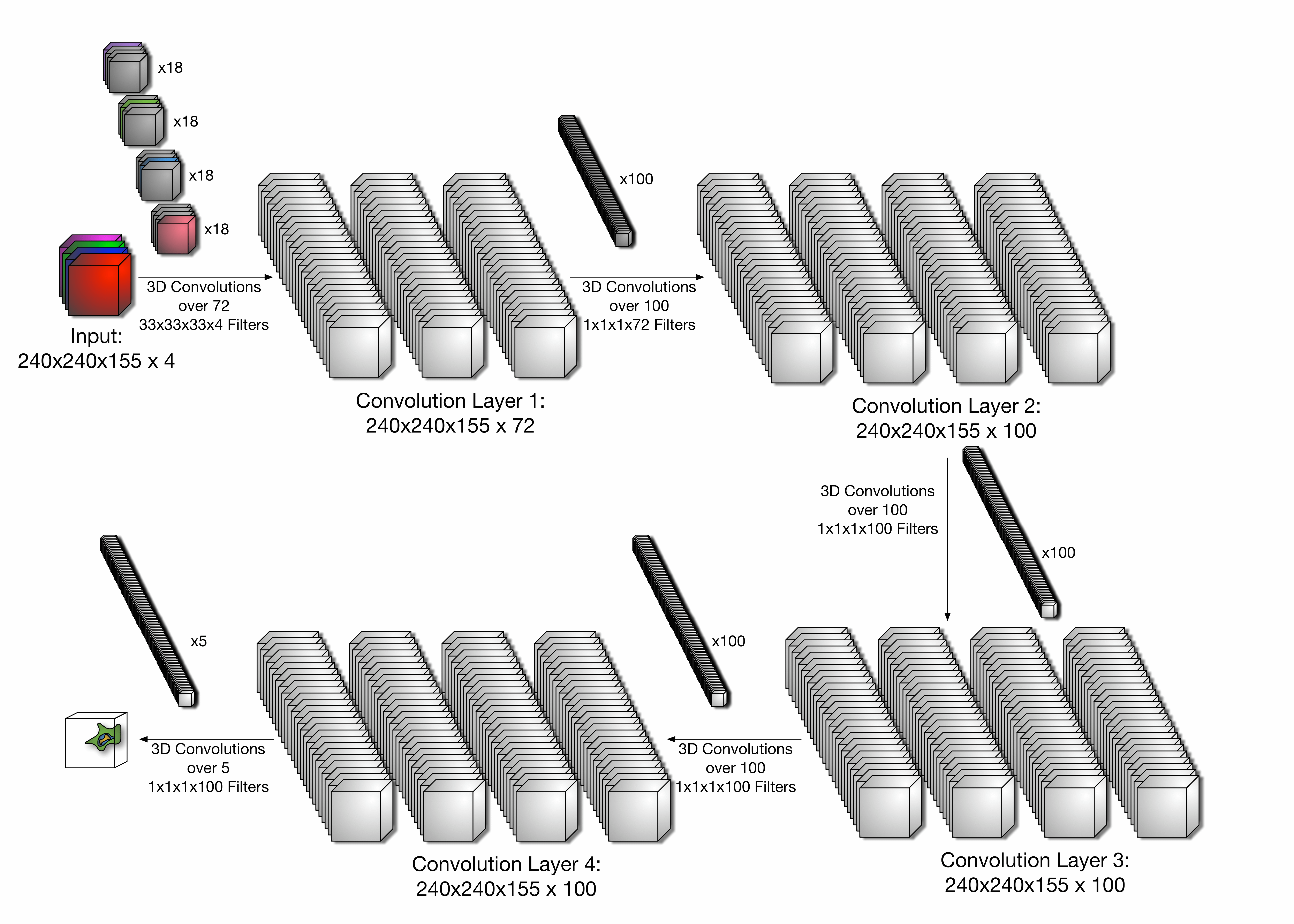}
	\caption{{3-D CNN architecture.} \small{Starting with 4 MRI modalities as input samples, we convolve 72 sparse 3-D Difference-of-Gaussian (DoG) filters to form the first convolutional layer.  The remaining convolutional layers use $1 \times 1 \times 1$ (scalar) filters over all channels to generate the subsequent layer, decoupling all pixels after the initial convolution.  The last layer produces 5 channels, each corresponding to a probability for classifying each pixel into non-tumor or 4 tumor subregions. }}
	\label{fig:methodologyConv}
\end{figure}

Our CNN architecture utilizes 5 convolutional layers (Figure $\ref{fig:methodologyConv}$).  Starting from $240 \times 240 \times 155$ inputs of volumetric image data in 4 MRI modalities (channels), the first layer performs 3-D convolution over all input channels with 72 pre-defined $33 \times 33 \times 33 \times 4$ filters. We then train 4 convolutional layers of $1 \times 1 \times 1$ filters over the number of channels in the preceding layer (72 for the first and 100 for all subsequent layers). The final output layer of 5 channels represents scores for the predicted class probabilities of each pixel as either non-tumor or 4 tumor subregions.

\subsection{3-D Convolution}
Next, we minimized the potential bias of the CNN architecture by considering the volumetric image data as a 3-D space of pixels. For 3-D image $I$ and filter $f$, the usual 2-D filters of CNNs can be generalized to 3-D convolutions (Eq.~$\ref{eq:3DConv}$) as defined below,
\begin{equation}\label{eq:3DConv}
(I * f) [x,y,z] = \sum_{\tau_x = 1}^{n_x} \sum_{\tau_y = 1}^{n_y} \sum_{\tau_z = 1}^{n_z}  I[\tau_x, \tau_y, \tau_z] \cdot f[x - \tau_x, y - \tau_y, z - \tau_z]
\end{equation} 
Given $n \times n \times n$ images and $m \times m \times m$ filters, time complexity of 3-D convolution is $\mathcal{O} (m^3 n^3)$. Since convolution in space is equivalent to element-wise multiplication in Fourier space, this complexity can be reduced to $\mathcal{O} \left(n^3 \log n \right)$. \\
\indent Previously, CNNs have repeatedly been found to have a first layer with trained weights that resemble Gabor-like filters~\cite{Yosinski14}.  Thus, to save computational time in training the model, we pre-select the first layer's filters to function as edge detectors. More specifically, we use 3-D Difference-of-Gaussian (DoG) filters, each represented by the difference of two normalized 3-D Gaussians of scales $\sqrt{2} \sigma$ and $\sigma$, as defined in Eq.~\ref{eq:3DDoG}.
\begin{equation}\label{eq:3DDoG}
\text{DoG} (\sigma) = \frac{1}{(2 \pi \sigma^2)^{3/2}} e^{-\frac{x^2 + y^2 + z^2}{2 \sigma^2}} - \frac{1}{(\pi \sigma^2)^{3/2}} e^{-\frac{x^2 + y^2 + z^2}{\sigma^2}}
\end{equation}
We created 8 filters of size $33 \times 33 \times 33$ with scales $\sigma = [\sqrt{2}, 2, 2 \sqrt{2}, \dots, 16]$.  Previous algorithms have shown the efficacy of DoG filters in blob detection \cite{sift}; in particular, their rotational symmetry enables the CNN to learn a blob profile for each pixel.  By contrast, while Gabor texture filters have emerged as a common theme in deep learning on image data \cite{Krizhevsky12}, their lack of rotational symmetry requires learning a full feature vector for each pixel for every possible orientation, which greatly increases learning complexity. \\
\indent Next, we apply the 8 DoG filters to the original input as well as the magnitude of the gradient of those images. Thus, we create 18 ``feature'' images: the original pixel intensities and their 8 filter products as well as the magnitude of the gradient values and their 8 filter products. After applying such computation on 4 MRI modalities, such design leads to a 72-dimensional feature space for each pixel. Overall, this non-trained convolutional layer results in a 3-D convolution of the $240 \times 240 \times 155 \times 4$ input data using 72 pre-defined $240 \times 240 \times 155 \times 4$ filters.

\subsection{Subsequent Convolution Layers as Pixel-wise Neural Network}
Each subsequent convolution layer consists of  $1 \times 1 \times 1$ kernels over all input channels. This choice enabling training on a CPU cluster is motivated by two benefits: (1) drastic decrease in the number of weights to be trained and (2) decoupling of pixels, allowing for a fully connected neural net implementation of the last five convolution layers. This decoupling is possible because the convolution layers and the softmax loss function operate independently for each pixel. \\
\indent Thus, given a 72-dimensional feature vector produced for each pixel by the first convolutional layer, we classify each pixel using a fully connected neural network. Our network architecture consists of 3 hidden layers of 100 neurons each, with a rectified linear unit (ReLU) as the activation function and the softmax function as the loss function.  The output layer of five neurons predicts the following classes: $0 = \text{non-tumor}$, $1 = \text{necrosis}$, $2 = \text{edema}$, $3 = \text{non-enhancing}$, and $4 = \text{enhancing}$.  The final classification step follows a voting algorithm as described previously for pooling expert segmentations from BRATS dataset \cite{brats}. 



\section{Experiments}
\subsection{Brain Tumor MRI Data}
\begin{figure}[htb]
	\centering
	\begin{subfigure}[h]{0.23\textwidth}
		\includegraphics[width=\textwidth]{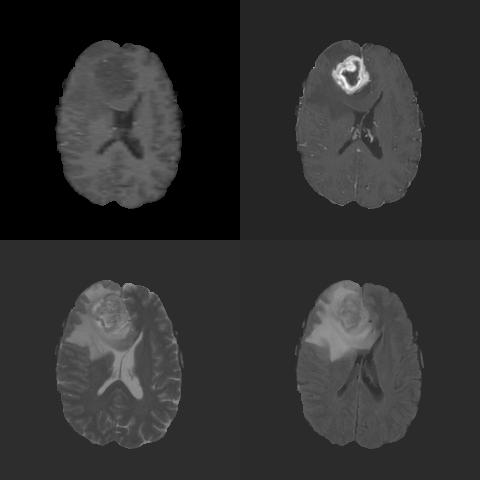}
		\caption{Modalities}
	\end{subfigure}
	~
	\begin{subfigure}[h]{0.23\textwidth}
		\includegraphics[width=\textwidth]{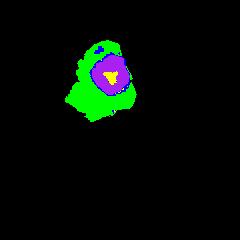}
		\caption{Labels}
	\end{subfigure}
	~
	\begin{subfigure}[h]{0.23\textwidth}
		\includegraphics[width=\textwidth]{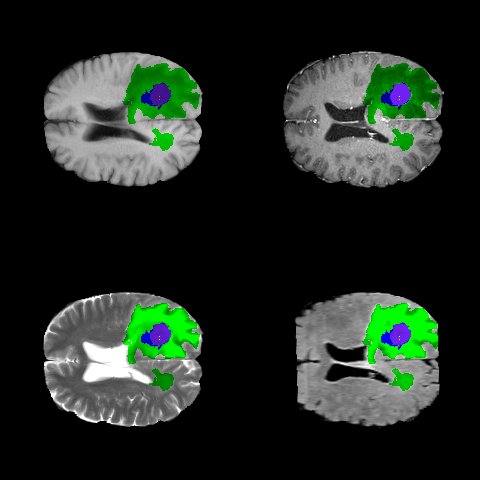}
		\caption{Modalities}
	\end{subfigure}
	\caption{{BRATS image data.} \small{(a) The four imaging modalities (upper left: T1-Pre, upper right: T1-Post, lower left: T2W, lower right: FLAIR).  (b) The four subregions (yellow: necrosis, green: edema, blue: non-enhancing, purple: enhancing), corresponding to the scans in (a).  (c)  A visualization of the labels superimposed on each modality.}}
	\label{fig:data}
\end{figure}

We used the Brain Tumor Image Segmentation Challenge (BRATS)\cite{brats} to evaluate performance of the proposed approach.  The 2015 BRATS data set consists of 274 samples: 220 patients with high-grade GBM (HGG) and 54 with low-grade GBM (LGG).  Each patient has 4 modalities (T1 post-contrast, T1 pre-contrast, T2-weighted, and FLAIR) and an expert segmentation that we treat as ground truth. The expert segmentation, which provides pixel-wise labeling into five segmentations based on the consensus of eleven radiologists: $0 = \text{non-tumor}$, $1 = \text{necrosis}$, $2 = \text{edema}$, $3 = \text{non-enhancing}$, and $4 = \text{enhancing}$. We additionally included BRATS 2013 dataset to compare with prior studies. All images were pre-processed by stripping the skull, co-registering images, and interpolating images to $240 \times 240 \times 155$ pixels (Figure \ref{fig:data}).

\subsection{Evaluation}
We evaluated our algorithms by focusing on three clinically relevant segmentations: ``whole'' or ``total'' referring to the entire tumor, ``core'' including all structures except ``edema,'' and ``active'' including only ``enhancing'' subregions unique to HGG \cite{brats}. For each of these three regions, accuracy is reported using the Dice coefficient by comparing the predicted segmentation  with the expert reference and with previously developed algorithms. This score, given in Eq.~$\ref{eq:dice}$, is equivalent to the harmonic mean of the precision and the recall.
\begin{equation}\label{eq:dice}
\text{Dice score} = \frac{2 \left| \text{Pred} \cap \text{Ref} \right|}{\left| \text{Pred} \right| + \left| \text{Ref} \right|},
\end{equation}

\section{Results}
\begin{figure}[htb]
	\centering
	\includegraphics[width = 0.9\textwidth]{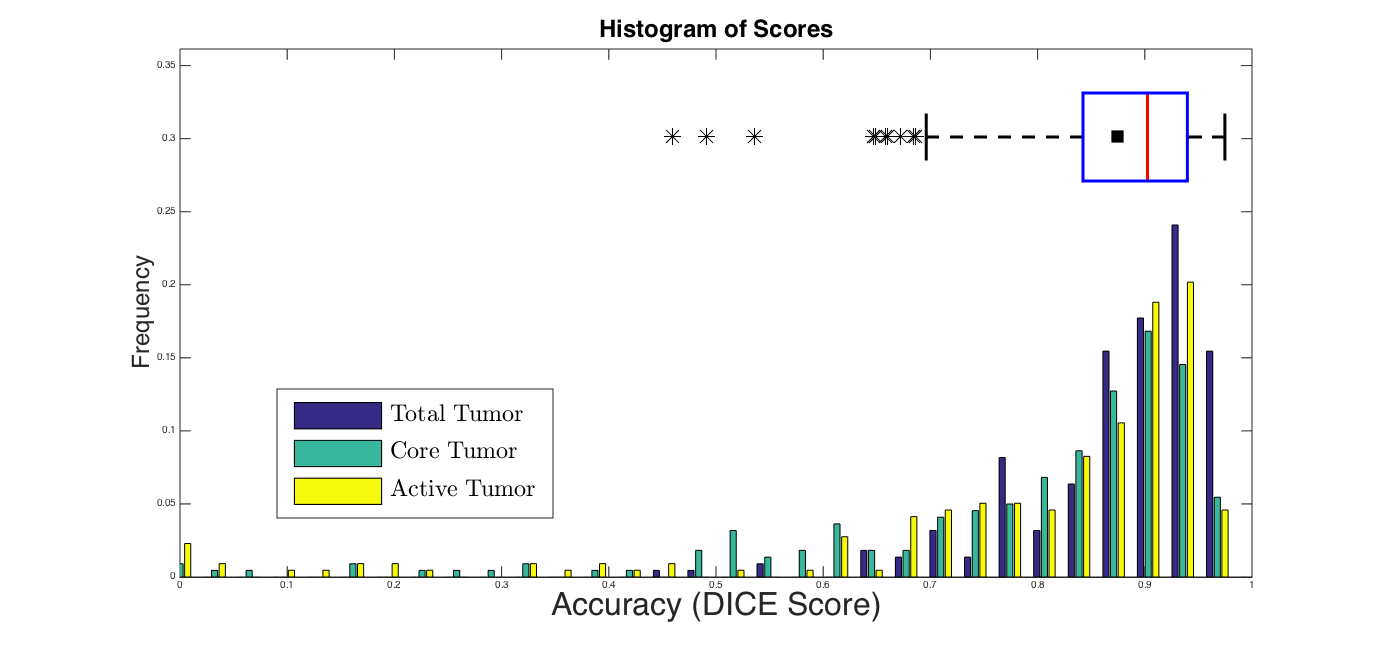}
	\caption{{Histogram of dice scores.} \small{The full distribution of dice scores for all 274 patients are shown for each of the three tumor regions (blue: total, turquoise: core, yellow: active). The box-and-whisker plot in the figure summarizes the distribution of accuracy for tumor segmentation.}}
	\label{fig:histogramAccuracy}
\end{figure}

\begin{figure}[htb]
	\centering
	\includegraphics[width=0.85\textwidth]{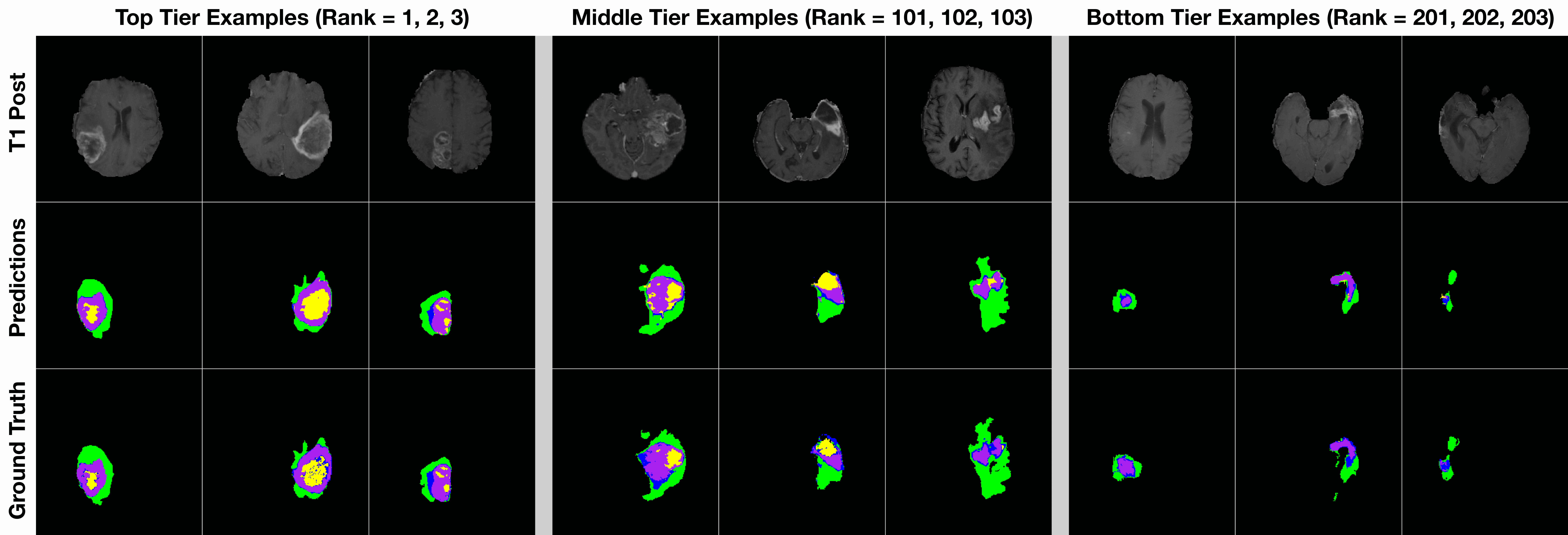}
	\caption{{Representative segmented slices.} \small{Results of the algorithm are given for three representative patients each with (left) high, (middle) intermediate, (right) and low Dice scores. Each column shows (top) the T1 post-contrast image, (middle) the predicted tumor subregion labeling, and (bottom) the expert segmentation labeling for a slice from that patient.  Label colors correspond to those of Figure \ref{fig:data}. \\}}
	\label{fig:results}
\end{figure}

The performance of the 3-D CNN  for ``total,'' ``core,'' and ``active'' tumor regions on the 2015 data set is shown (Figure $\ref{fig:histogramAccuracy}$), with a median accuracy of over $90\%$ on total tumor detection (compared to inter-radiologist reproducibility of $85\%$).  Slice-level comparison of our algorithm's labels with the expert segmentation is shown for representative samples of varying Dice scores (Figure \ref{fig:results}). \\
%
%
\indent Table \ref{table:Comparison} compares the performance of our algorithm to expert segmentations and competing methods for brain tumor segmentation. Overall, our algorithm shows quite competitive results comparing to prior approaches. First, comparing to raters, our results are comparable with annotations by individual radiologists and even close to results of expert segmentation generated by a voting algorithm~\cite{brats}. Additionally, we evaluated the performance of our method on the 2013 BRATS data set, comparing it to the best combination of programs from the 2013 BRATS challenge \cite{brats}. While each individual program from the 2013 challenge has lower performance than the combination, our algorithm, trained only on the 2013 data, has equal or better results than the combination in all three categories.  Finally, we compare our method with other methods on the 2015 data \cite{brats2015}. Our algorithm achieves Dice scores for whole, core, and active tumor detection of 87\%, 76\%, and 80\%, with the highest performance in two of the three clinically used regions. Our similar outcomes on both 2015 data and 2013 BRATS data reaffirmed the superior performance of the proposed 3-D CNN model with a notable improvement in classification accuracy of active tumor regions.

\begin{table}[h]
\caption{Performance Comparison (\%)}\label{table:Comparison}
\begin{center}
\begin{tabular}{L{2cm}|L{5cm}|L{2cm}|L{2cm}|C{1cm}}
\toprule
People&Description&Whole&Core&Active\\
& & \hspace*{\fill} HGG/LGG \normalsize & \hspace*{\fill} HGG/LGG& \\
\midrule
\midrule
Rater v. Rater & Comparison between radiologists using 2013 BRATS challenge data. & 85 \hfill (88/84) & 75 \hfill (95/67) & 74\\
Rater v. Fused & Comparison between radiologists and fused segmentation. & 91 \hfill (93/92) & 86 \hfill (96/80) & 85\\
Combination & The best combination of 2013 BRATS challenge programs using Algorithm 1. & 88 \hfill (89/86) & \textbf{78} \hfill (82 / 66) & \textbf{71} \\
\midrule
$\text{Ours}_{\text{2013}}$ & 3-D Convolutional Neural Network, using 2013 data set. & \textbf{89} \hfill (89/88) & \textbf{78} \hfill (79/74) & \textbf{71}\\
\midrule
\midrule
  Davy & Deep neural networks.  2014 Workshop. & 85 \hfill (-/-) & 74 \hfill (-/-) & 68 \\
Goetz & Extremely randomized trees.  2014 Workshop. & 83 \hfill (-/-) & 71 \hfill (-/-) & 68\\
Kleesiek & \emph{ilastik} Algorithm. 2014 Workshop. & 84 \hfill (84/82) & 68 \hfill (71/61) & 72\\
Kwon & GLISTR Algorithm.  2014 Workshop. & 88 \hfill (-/-) & \textbf{83} \hfill (-/-) & 72\\
Meier & Appearance and Context Sensitive Features.  2014 Workshop.  & 83 \hfill (84/-) & 66 \hfill (73/-) & 68\\
\midrule
$\text{Ours}_{\text{2015}}$ & 3-D Convolutional Neural Network, using 2015 data set. & \textbf{89} \hfill (89/87) & 76 \hfill (79/69) & \textbf{80}\\
\midrule
\midrule
\end{tabular}
\end{center}
\end{table}


\section{Conclusion}
We have proposed a 3-D fully convolutional netwroks that generalizes conventional CNNs in learning 3-D tumor MRI data. Specifically, we first use a non-trained convolutional layer with pre-defined DoG filters to perform true 3-D convolution that incorporates information about the local neighborhood at each pixel of the output. We then use three trained convolutional layers that act to decouple voxels, under the assumption that voxels are coupled only by the information already incorporated in the initial 3-D convolution. This architecture of a fully connected neural network at the level of pixels allows us to greatly increase the effective training data size from the number of patient samples to the number of pixels. We show that the use a modified non-trained convolutional layer can greatly reduces variance by increasing the number of training samples. It is known that patient-based samples can theoretically allow for complex features that relate wholly different parts of the brain, but the presented voxel-based training data allows the fully connected feed-forward neural network to learn higher-level features based on a much larger training data set in pixel space. Overall, our generalization to a 3-D CNN incorporates several key innovations addressing problems with existing approaches to using deep learning in medium-sized imaging data sets. 
%
%
%
%
%
%
%
%

\section{Acknowledgement}
This work was supported by the National Institutes of Health (NIH) under Award Number R01EB020527. 
\section{References}\label{refer}
\begingroup
\renewcommand{\section}[2]{}


\begin{thebibliography}{99}
\bibitem{Adamson09}
Adamson C, et al.  Glioblastoma multiforme: a review of where we have been and where we are going.  Expert Opin. Investig. Drugs {\bfseries 18}(8) (2009) 1061-1083.
\end{thebibliography}
\end{document}